\documentclass[conference]{IEEEtran}
\IEEEoverridecommandlockouts
\usepackage[square,sort,comma,numbers]{natbib}
\setcitestyle{citesep={,}}
\usepackage{lipsum, cuted}
\usepackage{amsmath,amssymb,amsfonts}
\usepackage{adjustbox}
\usepackage{stfloats}
\usepackage{algpseudocode}
\usepackage{graphicx,wrapfig}
\usepackage{multirow}
\usepackage{physics}
\usepackage[algo2e]{algorithm2e} 
\usepackage[utf8]{inputenc}
\usepackage{graphicx,wrapfig,lipsum}
\usepackage{listings}
\usepackage{float}
\usepackage{amsfonts}
\usepackage{algorithm}
\usepackage{algcompatible}
\usepackage{natbib}
\usepackage{color}   
\usepackage{hyperref}
\hypersetup{
    colorlinks=false, 
    linktoc=all,     
    linkcolor=blue,  
}
\usepackage{caption}

\usepackage{stmaryrd} 
\usepackage{amsmath}
\usepackage{amssymb}
\usepackage{amsfonts}
\usepackage{color}
\usepackage{textcomp}
\setcitestyle{square}
\usepackage{arydshln}

\usepackage{xcolor}
\def\BibTeX{{\rm B\kern-.05em{\sc i\kern-.025em b}\kern-.08em
    T\kern-.1667em\lower.7ex\hbox{E}\kern-.125emX}}

\begin{document}
\title{Knowledge-Based Convolutional Neural Network for the
Simulation and Prediction of Two-Phase Darcy Flows}
\author{\IEEEauthorblockN{Zakaria Elabid\textsuperscript{1}, Daniel Busby\textsuperscript{2}, Abdenour Hadid\textsuperscript{1}}
\IEEEauthorblockA{\textsuperscript{1} Sorbonne Center for Artificial Intelligence, Sorbonne University Abu Dhabi, UAE\\
\textsuperscript{2} TotalEnergies, France
}
}


\maketitle
\begin{abstract}
Physics-informed neural networks (PINNs) have gained significant prominence as a powerful tool in the field of scientific computing and simulations. Their ability to seamlessly integrate physical principles into deep learning architectures has revolutionized the approaches to solving complex problems in physics and engineering. However, a persistent challenge faced by mainstream PINNs lies in their handling of discontinuous input data, leading to inaccuracies in predictions.
This study addresses these challenges by incorporating the discretized forms of the governing equations into the PINN framework.
We propose to combine the power of neural networks with the dynamics imposed by the discretized differential equations. By discretizing the governing equations, the PINN learns to account for the discontinuities and accurately capture the underlying relationships between inputs and outputs, improving the accuracy compared to traditional interpolation techniques. Moreover, by leveraging the power of neural networks, the computational cost associated with numerical simulations is substantially reduced. We evaluate our model on a large-scale dataset for the prediction of pressure and saturation fields demonstrating high accuracies compared to non-physically aware models.
\end{abstract}
\section{Introduction}
Physics-informed neural networks (PINNs) \cite{OGphysics,PINN} have gained prominence due to their unique capabilities in modeling and predicting complex physical phenomena. One of the primary advantages of PINNs is their ability to seamlessly integrate the governing equations into the neural network architecture. A substantial challenge in applying PINNs to real-life dynamics applications lies in dealing with discontinuities and sharp gradients \cite{buckley-leverett}, often prevalent in hydrology and reservoir problems \cite{reservoir}.

In fluid dynamics problems \cite{flow}, predicting pressure and saturation traditionally relies on numerical simulations \cite{simulation}. These simulations employ complex mathematical models and computational techniques to approximate the behavior of fluids. While numerical simulations have been a cornerstone of fluid dynamics research, they do come with limitations. First, they often require substantial computational resources, making them computationally expensive for high-resolution simulations or real-time applications. Second, they demand a wealth of input data and, in many cases, idealized assumptions that may not fully capture the intricacies of real-world porous media \cite{3D-unet,3D-unet-Total}. Numerical simulations can struggle to handle discontinuities or sharp gradients in reservoir properties, and they may lack the flexibility to adapt to varying conditions and data scarcity. Additionally, the demand for extensive training data for deep neural networks can pose limitations in many porous media applications. Acquiring large datasets can be costly and time-intensive, necessitating strategies to overcome data scarcity while maintaining model accuracy. Moreover, the computational overhead associated with training deep neural networks, particularly for high-resolution simulations, presents a challenge, and renders the application of conventional neural networks methodologies highly impracticable.  While these shortcomings have spurred the exploration of alternative approaches like physics-informed neural networks (PINNs)  to address these challenges and enhance the accuracy and efficiency of fluid flow predictions in porous media\cite{reviewerSuggestion}, the abrupt changes in physical properties being pivotal for characterizing reservoir behavior made it imperative to develop techniques that enable PINNs to handle such discontinuities effectively \cite{uncertainty,KDL}. 

To overcome the shortcomings of the aforementioned techniques, we present a novel approach using discretized equations within the PINN to solve the two-phase flow problem in porous media. Our contributions in this paper are described as follows:
\begin{itemize}
    \item We build a physically aware generalizable framework called KED (Knowledge based encoder-decoder) where the network trains on observed data and the physical equations governing it to solve the problem of two-phase flow in a porous media.
    \item We compute the discrete spatio-temporal derivatives of the inputs and outputs to account for the discontinuity in the modeled data when enforcing the dynamics in the network.
    \item We evaluate our model on a large-scale dataset and demonstrate high accuracies of predictions compared to non physically aware models.
\end{itemize}

\section{Preliminaries}
This section discusses some necessary mathematical prelim-
inaries on the two-phase flow problem.
\subsection{Darcy's Law and Mass Conservation Equations}

Darcy's law \cite{Darcy} relates the fluid velocity to the pressure gradient in porous media and is fundamental to describing the flow of both oil and water through the porous medium. For fluid phase l, it can be expressed as:
\begin{equation}
q_l = -k \cdot \nabla P_l
\label{equation darcy}
\end{equation}

Where:
$q_l$ is the volumetric flow rate of fluid (water or oil), k is the permeability of the porous medium and 
$\nabla P_l$ is the pressure gradient of the fluid phase

For each fluid phase (oil and water), mass conservation equations describe how the fluid mass is distributed within the porous medium. These equations are typically expressed as:
\begin{equation}
\frac{\partial (\phi S_l \rho_l)}{\partial t} + \nabla \cdot (\rho_l \mathbf{q}_l) = Q_l
\label{equation darcy 2}
\end{equation}

Here, $\phi$ is porosity of the porous medium,
$S_l$  is the saturation of fluid,
$\rho_l$ is the density of fluid,
$q_l$ is flux rate of fluid,
$Q_l$ is sink term for fluid.

Solving these equations simultaneously in a porous medium with appropriate boundary conditions and initial conditions allows for the prediction of how oil and water distribute and flow through the porous media over time.

\begin{figure*}[bp]
\begin{equation}
\begin{aligned}
N={} & (\rho_l \frac{K k_{r l}}{\mu_l})_{\substack{i+\frac{1}{2}\\ j, k}} \frac{\hat{P}_{l i+1, j, k}^{n+1}-\hat{P}_{l i, j, k}^{n+1}}{d x^2}-(\rho_l \frac{K k_{r l}}{\mu_l})_{\substack{i-\frac{1}{2}\\ j, k}} \frac{\hat{P}_{l i, j, k}^{n+1}-\hat{P}_{l i-1, j, k}^{n+1}}{d x^2}+ 
 \rho_l (\frac{K k_{r l}}{\mu_l})_{\substack{i, j+\frac{1}{2}\\ k}} \frac{\hat{P}_{l i, j+1, k}^{n+1}-\hat{P}_{l i, j, k}^{n+1}}{d y^2}\\ &-(\rho_l \frac{K k_{r l}}{\mu_l})_{\substack{i, j-\frac{1}{2}\\ k}} \frac{\hat{P}_{l i, j, k}^{n+1}-\hat{P}_{l i, j-1, k}^{n+1}}{d y^2} +
(\rho_l \frac{K k_{r l}}{\mu_l})_{\substack{i, j\\ k+\frac{1}{2}}} \frac{\hat{P}_{l i, j, k+1}^{n+1}-\hat{P}_{l i, j,k}^{n+1}}{d z^2}-\left(\rho_l \frac{K k_{r l}}{\mu_l}\right)_{\substack{i, j\\ k-\frac{1}{2}}} \frac{\hat{P}_{l i, j,k}^{n+1}-\hat{P}_{l i, j, k-1}^{n+1}}{d z^2}\\ &+
 q_{l i, j,k}-\phi_{i, j,k} C_l \hat{S}_{l i, j,k} \rho_l \frac{\hat{P}_{l i, j,k}^{n+1}-\hat{P}_{l i, j,k}^n}{d t}-\phi_{i, j,k} \rho_l \frac{\hat{S}_{l i, j,k}^{n+1}-\hat{S}_{l i, j,k}^n}{d t}=0 \\  
\end{aligned} 
\tag{4}\label{equation discrete}
\end{equation}
\end{figure*}


\section{Proposed Method}
We consider a 3D two-phase flow problem in a porous media. The governing equations write:
\begin{equation}
\begin{aligned}
{}&\frac{\partial}{\partial x}\left(\rho_l \frac{k_x k_{r l}}{\mu_l} \frac{\partial P_l}{\partial x}\right)+\frac{\partial}{\partial y}\left(\rho_l \frac{k_y k_{r l}}{\mu_l} \frac{\partial P_l}{\partial y}\right)+\\ &\frac{\partial}{\partial z}\left(\rho_l \frac{k_z k_{r l}}{\mu_l} \frac{\partial P_l}{\partial z}\right)  +q_l=\phi C_l S_l \rho_l \frac{\partial P_l}{\partial t}+\phi \rho_l \frac{\partial S_l}{\partial t}    
\end{aligned}
\tag{3}\label{equation 1}
\end{equation}

Where: l is the fluid phase (oil or water), parameters $C,\rho,\mu,\phi,k_r$ are known for each phase, $k_x = k_y = k_z = K$ is the input, and P and S are the pressure and the saturation.

From the differential equation, we simulate an experiment using a built-in simulation tool to generate 3D instances of permeability K (input) and their corresponding pressure maps P and saturation maps S.

We solve the discrete form of equation \ref{equation 1} by using Finite Differences method \cite{FD}
which we will be using as the governing equation of our proposed PINN framework.

The previous equations can be rewritten into equation \ref{equation discrete} for steps $i,j,k$ in space corresponding to the grid positions and n in time. $d t,d x,d y,d z$ are the lengths of the time step and spacial steps in directions x,y and z respectively. The notation $i+\frac{1}{2}$ and $i-\frac{1}{2}$ corresponds to the harmonic mean. It's given for i (and similarly for j and k) by the expression:
\begin{equation*}
   k_{i+\frac{1}{2}\\ j, k} = \frac{2k_{i,j,k}k_{i+1,j,k}}{k_{i,j,k}+k_{i+1,j,k}} ; k_{i-\frac{1}{2}\\ j, k} = \frac{2k_{i,j,k}k_{i-1,j,k}}{k_{i,j,k}+k_{i-1,j,k}}
\end{equation*}

Our model consists of two identical encoder-decoder networks one to predict pressure maps and the other to predict saturation maps. We train our model in two different manners:
\subsubsection{Supervised Learning}
The model predicts pressure and saturation maps from permeability fields K and timesteps input t. 
The resulting maps are then compared to the real maps using prediction metrics such as MAE or MSE:
\begin{center}
$L_{P}=MSE(P_{predicted}, P_{real})$

$L_{S}=MSE(S_{predicted}, S_{real})$    
\end{center}

\subsubsection{Knowledge-Based Learning}

To compute the discrete derivatives from equation \ref{equation discrete}, we consider that the values of \(P(x, y, z, t),S(x, y, z, t)\) represent the output corresponding to inputs \(K(x, y, z)\) and \(t\), while \(P(x + dx, y, z, t),S(x + dx, y, z, t)\) correspond to \(K(x + dx, y, z)\) and \(t\), and \(P(x, y, z, t + dt),S(x, y, z, t + dt)\) correspond to \(K(x, y, z)\) and \(t + dt\), and so forth.


We train our networks in a semi-supervised \cite{semi-supervised} manner by satisfying the underlying physics in equation \ref{equation discrete}
for every gridcell $i,j,k$ and every timestep t of the duration of the simulation.

The corresponding physics loss writes:
\begin{equation}
    L_{phy}=\sum_{t=0}^{t_n}\sum_{i=1}^{N_x/dx}\sum_{j=1}^{N_y/dy}\sum_{k=1}^{N_z/dz} N(i,j,k,K,t)
    \label{physical loss}
\end{equation}
where N is the differential operator described by equation \ref{equation discrete} and $N_x,N_y,N_z$ is the length of the grid blocks in the directions $x,y,z$ each block of dimensions $dx,dy,dz$.

An algorithmic version of the suggested model is presented in Algorithm \ref{KDL_algo}.

\subsection*{U-Net Encoder Decoder}

U-Net architecture is an encoder-decoder structure, characterized by a contracting path to capture context and a symmetry enabling precise localization. It is especially adept at handling multi-dimensional subsurface flow data. The U-Net architecture learns effectively complex mappings from input 3D permeability  to predict the dynamic evolution  of the output pressure and saturation maps. For our method, we use a simple U-Net composed of 4 convolutional layers with a kernel of 3 and a stride of 1 for the encoder, a linear layer with 100 neurons and 4 deconvolution layers with a kernel of 3 and stride of 1 for the decoder. Encoder and decoder layers are linked through skip-connections to preserve local physical properties.

\begin{algorithm}
\caption{\textsc{Knowledge-Based Encoder-Decoder (KED)}}\label{KDL_algo}
\KwData{X = Permeability Field $K(x,y,z)$, Virtual Permeability field $K_v(x,y,z)$, Time Matrix $[t_1,...,t_n]$}

\textbf{Networks:} Net1 and Net2 are identical encoder-decoders

\KwResult{Y = Pressure field $P(x,y,z,t)$, Saturation field $S(x,y,z,t)$}

Initialize networks' parameters: (weights and biases)\;

\While{$epoch<max\_epochs$ \textbf{or} Termination Condition} {
Load pretrained network state (weights, biases)\;

Compute $P(t) = Net1(k,t)$,$S(t) = Net2(k,t)$\;

Calculate the loss $L_{P} = MSE(P_{predicted}, P_{real})$\;

Calculate the loss $L_{S} = MSE(S_{predicted}, S_{real})$\;

 Backpropagate $\lambda_pL_{P} +\lambda_sL_{S} $ (using ADAM optimizer) $\lambda_p,\lambda_s$ being tuning parameters chosen by cross-validation\;
}
Load network state

\While{$epoch<max\_epochs$ \textbf{or} Termination Condition}{
Compute $P_v(t) = Net1(k_v,t)$, $S_v(t) = Net2(k_v,t)$\;

Compute $P_v(t+dt) = Net1(k_v,t+dt)$, $S_v(t+dt) = Net2(k_v,t+dt)$\;

Calculate the loss $L_{phy}$ from Equation \eqref{equation discrete} using the discrete forms $P_v(t),P_v(t+dt),S_v(t),S_v(t+dt),k_v,t,dt$ to compute the discrete derivatives $\forall t$;

Backpropagate $\lambda_{phy} L_{phy}$ (using ADAM optimizer) $\lambda_{phy}$ being a tuning parameter chosen by cross-validation\;

}

\end{algorithm}
\section{Experimental Analysis}

\begin{wrapfigure}[12]{r}{0.2\textwidth}
\centering

\includegraphics[width=0.18\textwidth]{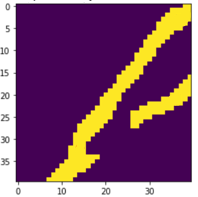}
\caption{Permeability map with sand and mud facies}
\label{permeability}
\end{wrapfigure} 

\textit{A. Dataset}

\vspace{0.1cm}
We evaluate our KED approach on a large dataset containing 2923 realizations of permability K fields corresponding to two facies (as can be seen in Figure \ref{permeability}): sand and mud. For the sand phase (yellow), K=2000md and for the mud phase (purple), K=20md. Similarly, the porosity $\phi=0.25$ for the sand and $\phi=0.1$ for the mud. The dataset contains 40m x 40m x 20m grid blocks - each block being of dimensions dx = 20, dy = 20, dz = 2 \cite{Data_assimilation}. The simulation is done over 21 timesteps of 50 days each. For this problem, the set values of parameters are $C = 9.10^{-5}$, $\rho_{water} = 1838 Kg/m^3 $, $\rho_{oil} = 787 Kg/m^3$, $\mu_{water}=0.31cp$, $\mu_{oil}=1.14cp$.

For simplification, the problem considers that both water and oil undergo similar pressure levels $P_o = P_w = P$ and that oil and water saturations are complementary i.e $S_o = 1-S_w $ which means that there can only be oil or water in the reservoir and the presence of any other fluid is negligible. Additionally, the relative permeability $k_r$ ($k_{ro}$ for oil and  $k_{rw}$ for water) can be expressed as a function of water saturation $S_w$ \cite{Relative-perm}:

$$k_{r o}\left(S_w\right)=k_{o}\left(\frac{1-S_w-S_{c}}{1-2S_{c}}\right)^a; k_{r w}\left(S_w\right)=k_{w}\left(\frac{S_w-S_{c}}{1-2S_{c}}\right)^a$$

Where: $a=2$, $S_{c}$ = 0.1, $k_o$ =1 and $k_w$ = 0.8

The data is publicly available at \cite{data}.
\setcounter{subsection}{1}
\subsection{Performance measures}
To evaluate our approach, we consider two metrics commonly used in prediction models. A better model corresponds to lower values of these metrics : (a) Mean Absolute Error (MAE) and (b) Mean Squared Error (MSE). Nonetheless, in cases of a homogeneous noise free input data, these metrics are bound to be low even at worst case scenario. Therefore, to assess the performance of the network in capturing the underlying physics of the problem, we consider an additional physically relevant metric. This metric corresponds to the amount of oil production of the reservoir throughout the simulation process in each production well. This metric can be computed using Peaceman equation\cite{peaceman} :

\begin{equation}
     q_w=\frac{2\pi k d z}{ln(\frac{r_0}{r_w})}\frac{k_r\rho}{\mu}(P-bhp)
     \label{peaceman}
\end{equation}

where $r_0 = 0.14\sqrt{d x^2+d y^2}$,  $r_w = 0.1$ and $bhp$ is the bottom hole pressure fixed at 310 bar for our experiment.


\subsection{Implementation of the proposed framework}
Data is acquired through numerical simulation of physical equations in Section II-A using AD-GPRS simulator \cite{datagenerator}. The code is written in pytorch workflow.
We train two identical encoders-decoders with skip-connections in a supervised manner on realizations of permeability. We preprocess the permeability into a matrix containing the timesteps as well. We feed the batch to the $1^{st}$ network to output the 3D pressure field in time P(x,y,z,t) and the $2^{nd}$ network to output the 3D saturation field in time S(x,y,z,t). We use MSE as a loss for the backpropagation. After loading the network state \cite{transfer}, we train our network in a semi-supervised manner where from additional realizations of permeability k, which we call virtual permeability in Algorithm \ref{KDL_algo} to mitigate potential ambiguity with the realizations used in supervised training, we compute  the pressure and saturation fields and their discrete derivatives as described by Equation \ref{equation discrete}. The feedback mechanism consists on minimising the observed loss (identical to pre-training) and physical loss as expressed in Equation \ref{physical loss}. The total loss can be expressed as $L_{total}=L_{data}+\lambda L_{physical}$
where $\lambda=0.1$

\subsection{Visualization of 2D pressure and saturation fields}
\begin{figure}
    \centering
c    \includegraphics[width = 0.47\textwidth]{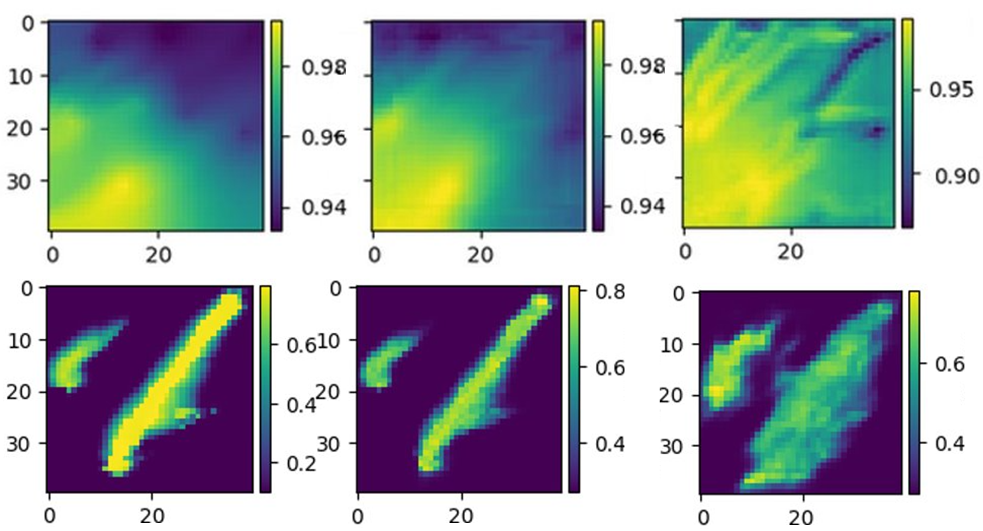}    
\captionsetup{belowskip=-15pt}
\caption{Comparison between reference (left), KED (middle) and non-physics prediction baseline (right) of 2D Pressure maps (top) and saturation maps (bottom)}
    \label{fig:saturation}
\end{figure}

Figure \ref{fig:saturation} displays the 2D predictions of pressure saturation maps at specific depths using both our approach (middle) and the same encoder-decoder model without the inclusion of dynamics (right). We can notice that our network manages to make accurate and physically consistent predictions compared to the baseline non-physically aware model. The shortcomings of non-physics models are particularly apparent in saturation maps where oil appears flow at higher rates in the baseline prediction compared to the reference, potentially leading to inconsistencies in the well drilling process. This shortcoming is avoided with our knowledge-based model that manages to properly captures the feature maps of pressure and satutration due to its awareness of the discontinuities of the dynamics.
\subsection{Production values}
\begin{figure}
\centering
\includegraphics[width = 0.47\textwidth]{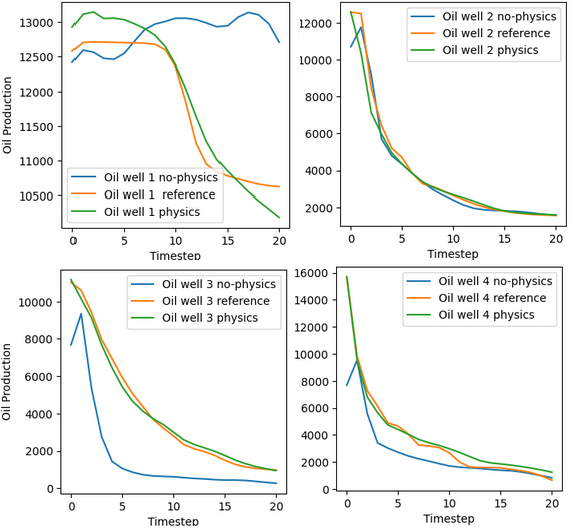} 
\caption{Comparison of oil production in production wells between reference (orange), baseline (blue) and our approach (green)}
\label{fig:wells}
\end{figure}

\begin{figure}
    \centering
    \includegraphics[width = 0.45\textwidth]{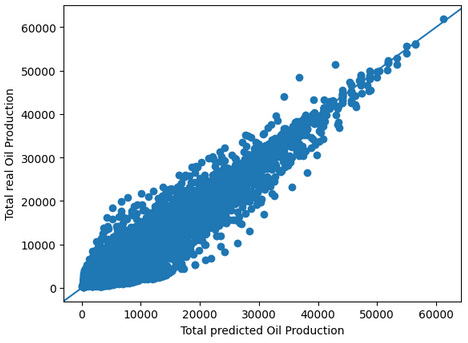}
     \captionsetup{belowskip=-15pt}
    \caption{Crossplot of the real oil production and the predicted oil production levels using our approach on 500 test samples}
    \label{fig:production}
\end{figure}

The production values are computed using Equation \ref{peaceman}. In this simulation, there are 4 producing wells. The oil production of the wells is depicted in Figure \ref{fig:wells}. Based on the results obtained, the physically based encoder-decoder manages to predict production values closer to the real ones in addition to exhibiting the same behaviour as the reference production curve for all 4 wells at all timestamps. The disparity is  particularly apparent in this example at well 1 where the non-physically aware baseline model is uncapable of mimicking the  evolution curve of the groundtruth production, as well as the initial conditions where it exhibits unwanted behaviour, potentially related to the discontinuities in the system and practically leading to unquantifiable uncertainty in real life applications. The power of our physically-aware model can be further seen in crossplot of Figure \ref{fig:production} where we score closer production values to the real production.




\section{Conclusion}
In this paper, we presented a novel approach to physics-informed machine learning by combining physical knowledge in the form of discrete differential equations and observable knowledge from the data to solve two-phase flow problem in porous media. The proposed model takes a semi supervised approch in which it learns partially from the data, the discrete form of the dynamics to penalize the loss in addition to prior knowledge gained in supervised pre-training. We built our model under the assumption that the continuity of the derivatives in classic PINNs can overlap with the discontinuity of the reservoir grids. This hypothesis holds strong when comparing the performance of our network to baseline models. An interesting perspective is to try the proposed KED model with different architectures such as Generative Adversarial Networks (GAN) \cite{Gan} to further see the impact of physical knowledge on a network renowned for modelling and forecasting complex 3D feature maps. To support the principle of reproducible research and to allow a fair comparison, the source code of this work will be made available at a later date.
\section*{ACKNOWLEDGMENT}
The support of TotalEnergies is fully acknowledged. Zakaria ELabid (PhD Student) and Abdenour Hadid (Professor, Industry Chair at SCAI Center of Abu Dhabi) are funded by TotalEnergies collaboration agreement with Sorbonne University Abu Dhabi.

\clearpage
\bibliographystyle{IEEEtran}
\bibliography{biblio}
\end{document}